\def\year{2020}\relax
\documentclass[letterpaper]{article} 
\usepackage{aaai20}  
\usepackage{times}  
\usepackage{helvet} 
\usepackage{courier}  
\usepackage[hyphens]{url}  
\usepackage{graphicx} 
\usepackage{amsmath}
\usepackage{multirow}
\usepackage{graphicx}  
\title{Knowledge Integration Networks for Action Recognition}
\author{ Shiwen Zhang\textsuperscript{\rm 1,2} \quad Sheng Guo\textsuperscript{\rm 1,2} \quad Limin Wang\textsuperscript{\rm 3} \quad Weilin Huang\thanks{Weilin Huang is the corresponding author.}\textsuperscript{\rm 1,2,}  \quad Matthew R. Scott\textsuperscript{\rm 1,2} \\
\textsuperscript{\rm 1} Malong Technologies, Shenzhen, China \\
\textsuperscript{\rm 2} Shenzhen Malong Artificial Intelligence Research Center, Shenzhen, China \\
\textsuperscript{\rm 3} State Key Lab for Novel Software Technology, Nanjing University, China \\
}

\begin{document}

\maketitle

\begin{abstract}
In this work, we propose Knowledge Integration Networks (referred as KINet) for video action recognition. KINet is capable of aggregating meaningful context features which are of great importance to identifying an action, such as human information and scene context.
We design a three-branch architecture consisting of a main branch for action recognition, and two auxiliary branches for human parsing and scene recognition which allow the model to encode the knowledge of human and scene for action recognition.  We explore two pre-trained models as teacher networks to distill the knowledge of human and scene for training the auxiliary tasks of KINet.
Furthermore, we propose a two-level knowledge encoding mechanism which contains a Cross Branch Integration (CBI) module for encoding the auxiliary knowledge into medium-level convolutional features, and an Action Knowledge Graph (AKG) for effectively fusing high-level context information. This results in an end-to-end trainable framework where the three tasks can be trained collaboratively, allowing the model to compute strong context knowledge efficiently.
The proposed KINet achieves the state-of-the-art performance on a large-scale action recognition benchmark Kinetics-400, with a top-1 accuracy of 77.8\%.  We further demonstrate that our KINet has strong capability by transferring the Kinetics-trained model to UCF-101, where it obtains 97.8\% top-1 accuracy.

\end{abstract}

\section{Introduction}
\begin{figure}[t]
\centering
\includegraphics[width=1.0\columnwidth]{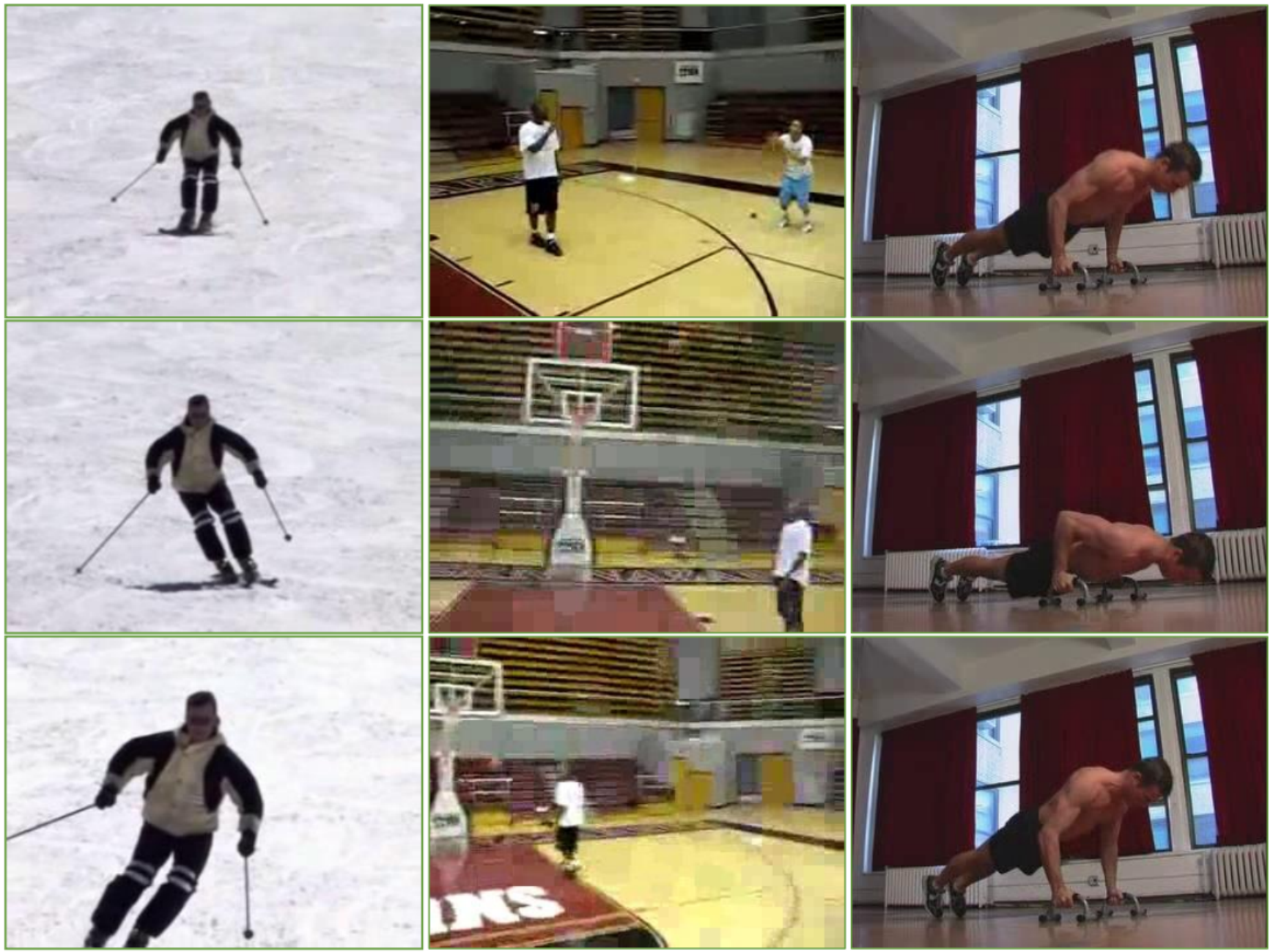} 
\caption{Human action is a high-level concept, which can be identified using various semantic cues, such as human and scene knowledge.}
\label{introduction}
\end{figure}

Deep learning technologies have recently advanced various tasks on video understanding, particularly for human action recognition where the performance has been improved considerably \cite{DBLP:conf/eccv/WangXW0LTG16,DBLP:conf/cvpr/CarreiraZ17,DBLP:conf/cvpr/0004GGH18}.
Intuitively, human action is a highly-semantic concept that contains various semantic cues. For example, as shown in Figure \ref{introduction}, in the first column, one can identify a skiing action by learning from a snow field scene with the related human dresses. In the second column, people can still know the action as basketball playing by observing basketball court and players, even it may be difficult to identify the ball due to low resolution or motion blur. In the last column, we can easily recognize a pushup action from human pose presented. Therefore, context knowledge is critical to understand human actions in videos, and learning meaningful such context knowledge is of great importance to improving the performance.

Past work commonly considered action recognition as  a classification problem, and attempted to learn action-related semantic cues directly from training videos \cite{DBLP:journals/corr/abs-1812-03982,DBLP:conf/cvpr/WangL0G18,DBLP:conf/cvpr/CarreiraZ17}. They assumed that action-related features can be implicitly learned with powerful CNN models by simply using video-level action labels.
However, it has been proven that learning action and actor segmentation jointly can boost the performance of both tasks. Experiments were conducted on A2D dataset \cite{DBLP:conf/cvpr/XuHXC15}, where the ground truth of actor masks and per-pixel action labels were provided.
Yet in practice, it is highly expensive to provide pixel-wise action labels for a large-scale video dataset, and such per-pixel annotations are not available in most action recognition benchmarks, such as Kinetics \cite{DBLP:conf/cvpr/CarreiraZ17} and UCF-101 \cite{DBLP:journals/corr/abs-1212-0402}.

Deep learning methods have achieved expressive performance on various vision tasks, such as human parsing \cite{DBLP:conf/cvpr/GongLZSL17}, pose estimation \cite{wang2019geometric}, semantic segmentation \cite{DBLP:conf/cvpr/ZhaoSQWJ17},
and scene recognition \cite{zhou2017places,wang2017knowledge}. It is interesting to utilize these existing technologies to enhance the model capability by learning context knowledge from action videos. This inspired us to design a knowledge distillation \cite{DBLP:journals/corr/HintonVD15} mechanism to learn the context knowledge of human and scene explicitly, by training action recognition jointly with  human parsing and scene recognition.
This allows the three tasks to work collaboratively, providing a more principled approach that learns rich context information for action recognition without additional manual annotations.

\textbf{Contributions.} In this work, we propose Knowledge Integration Networks, referred as KINet, for video action recognition. KINet is capable of aggregating meaningful
context features by design new three-branch networks with knowledge distillation. The main contribution of the paper is summarized as follows.

\begin{itemize}
\item We propose KINet - a three-branch architecture for action recognition. KINet has a main branch for action recognition, and two auxiliary branches for human parsing and scene recognition which encourage the model to learn the knowledge of human and scene via knowledge distillation. This results in an end-to-end trainable framework where the three tasks can be trained collaboratively and efficiently, allowing the model to learn the context knowledge explicitly.

\item We design a two-level knowledge encoding mechanism. The auxiliary human and scene knowledge can be encoded into convolutional features directly by introducing a new Cross Branch Integration (CBI) module. An Action Knowledge Graph (AKG) is further designed for effectively modeling the high-level correlations between action and the auxiliary knowledge of human and scene.

\item With the enhanced context knowledge of human and scene, the proposed KINet obtains the state-of-the-art performance with a top-1 accuracy of 77.8\% on Kinetics-400 \cite{DBLP:conf/cvpr/CarreiraZ17}, and also demonstrates strong transfer ability to UCF-101 dataset \cite{DBLP:journals/corr/abs-1212-0402}, by achieving a top-1 accuracy of 97.8\%.

\end{itemize}

\section{Related Work}

We briefly review recent work on video action recognition and action recognition with external knowledge.

Various CNN architectures have been developed for action recognition, which can be roughly categorized into three groups.
 First, a two-stream architecture was introduced in  \cite{Simonyan2014TwoStreamCN}, where one stream is used for learning from RGB images, and the other one is applied for modeling optical flow. The results produced by the two CNN streams are then fused at later stages, yielding the final prediction. Two-stream CNNs and its extensions have achieved impressive results on various video recognition tasks.
Second, 3D CNNs have recently been proposed in \cite{Tran2015LearningSF,DBLP:conf/cvpr/CarreiraZ17}, by considering a video as a stack of frames. The spatio-temporal features of action can be learned by 3D convolution kernels. However, 3D CNNs often explore a larger number of model parameters, which require more training data to achieve high performance. Recent results on Kinetics dataset, as reported in \cite{xie2018rethinking,DBLP:conf/cvpr/0004GGH18,DBLP:journals/corr/abs-1812-03982}, show that 3D CNNs can obtain competitive performance on action recognition.
 Third, recurrent networks, such as LSTM  \cite{Ng2015BeyondSS,Donahue2015LongTermRC}, have been explored for temporal modeling, where a video is considered as a temporal sequence of 2d frames. .

Recently, 
Xu et al. created an A2D dataset \cite{DBLP:conf/cvpr/XuHXC15}, where pixel-wise annotations on actors and actions were provided. In \cite{DBLP:conf/cvpr/XuC16}, the authors attempted to handle a similar problem by using probabilistic graphical models. A number of deep learning approaches have been developed for actor-action learning on the A2D databaset \cite{DBLP:conf/iccv/KalogeitonWFS17,DBLP:conf/bmvc/DangZ0HDY18,DBLP:conf/eccv/JiBSN18}, demonstrating that jointly learning actor and action representations can improve action recognition and understanding. However, these approaches are built on dense pixel-wise labelling on actors and actions provided in the A2D dataset, which are highly expensive and are difficult to obtain from a large-scale action recognition dataset, such as Kinetics \cite{DBLP:conf/cvpr/CarreiraZ17} and UCF-101 \cite{DBLP:journals/corr/abs-1212-0402}, where only video-level action labels are provided.

\begin{figure*}[t]
\centering
\includegraphics[width=0.75\textwidth]{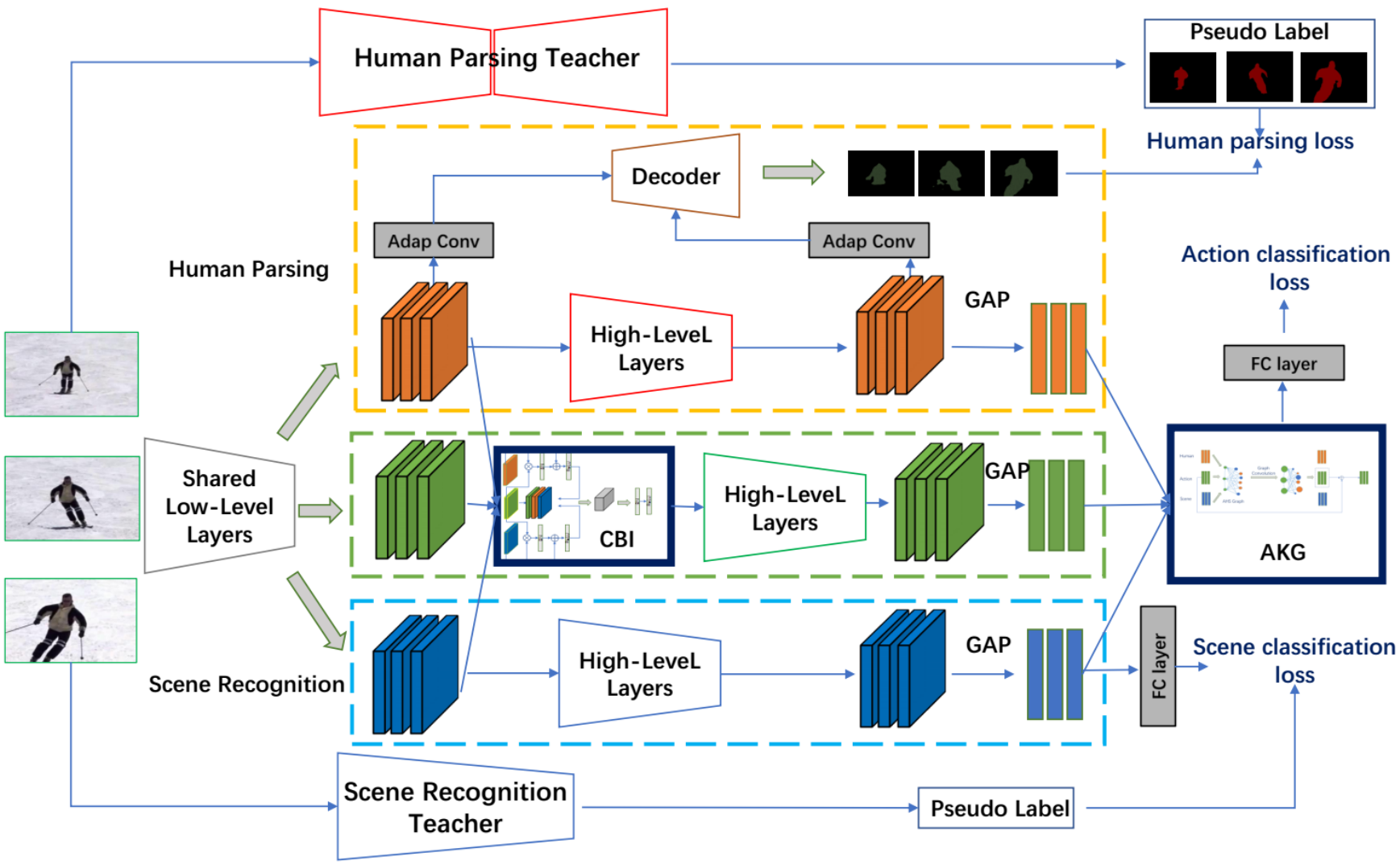} 
\caption{The overall structure of Knowledge Integration Networks for video action recognition. Details of CBI Module and AKG will be explained in following sections. Better viewed in color.}
\label{KInet}
\end{figure*}

An action can be defined by multiple elements, features or context information, and it is still an open problem on how to effectively incorporate various external knowledge into action recognition. In \cite{DBLP:conf/eccv/Ikizler-CinbisS10}, the authors tried to combine object, scene and action recognition by using multiple instance learning framework. Recently, with deep learning approaches, Jain et al. introduced object features to action recognition by discovering the relations of action and objects \cite{DBLP:conf/eccv/Ikizler-CinbisS10}, and Wu et al. further explored the relation of object, scene and action by designing  a more advanced and robust discriminative classifier \cite{DBLP:conf/cvpr/WuFJS16}.
In \cite{DBLP:conf/bmvc/WangSWGH16}, an external object detector was used to provide locations of objects and persons, and the authors in \cite{DBLP:conf/eccv/WangG18} incorporated an object detection network to provide object proposals, which are used with spatio-temporal graph convolutional networks for action recognition.

However, all these methods commonly rely on external networks to extract semantic cues, while such external networks were trained independently, and were fixed when applied to action recognition. This inevitably limits their capability for learning meaningful action representation. Our method is able to learn the additional knowledge of human and scene via knowledge distillation, allowing us to learn action recognition jointly with human praising and scene recognition with a signle model, providing a more efficient manner to encode the context knowledge for action recognition.

\section{Knowledge Integration Networks}

In this section, we describe details of the proposed {\bf K}nowledge {\bf I}ntegration {\bf N}etworks (KINet) which is able to distill the knowledge of human and scene from two teacher networks. KINet contains three branches, and has a new knowledge integration mechanism by introducing a Cross Branch Integration (CBI) module for encoding the auxiliary knowledge into the intermediate convolutional features, and an Action Knowledge Graph (AKG) for effectively integrating high-level context information.\\

\noindent\textbf{Overview.} The proposed KINet aims to explicitly incorporate scene context and human knowledge into human action recognition, while the annotations of scene categories or human masks are highly expensive, and thus are not available in many existing action recognition datasets. In this work, we attempt to tackle this problem by introducing two external teacher networks able to distill the extra knowledge of human and scene, providing additional supervision for KINet. This allows KINet to learn action, scene and human concepts simultaneously, and enables the explicit learning of multiple semantic concepts without additional manual annotations.

Specifically, as shown in Figure \ref{KInet}, KINet employs two external teacher networks to guide the main network. The two teacher networks aim at providing pseudo ground truth for scene recognition and human parsing. The main network is composed of three branches, where the fundamental branch is used for action recognition. The other two branches are designed for two auxiliary tasks - scene recognition and human representation, respectively. The intermediate representations of three branches are integrated by the proposed CBI module. Finally, we aggregate the action, human and scene features by designing an Action Knowledge Graph (AKG). The three branches are jointly optimized during training, allowing us to directly encode the knowledge of human and scene into KINet for action recognition.

We also tried to incorporate an object branch by distillation, but this leads to a performance drop with unstable results. We conjecture that there may be two main reasons: (1) Due to low resolution and motion blur, it is very hard to identify the exact object in the video. (2) The categories of object detection/segmentation are usually rather limited, while for action recognition, the objects involved are much more diverse. Although object proposals are used by \cite{DBLP:conf/eccv/WangG18} to overcome these problems, it is very hard to distill these noisy proposals. So we just use the ImageNet \cite{DBLP:conf/cvpr/DengDSLL009} pretrained model to initialize the framework instead of forcing it to "remember" everything from it.

\subsection{The Teacher Networks}
We explore two teacher networks pre-trained for human parsing and scene recognition.

{\bf Human parsing network}. We use LIP \cite{DBLP:conf/cvpr/GongLZSL17} dataset to train the human parsing network. LIP is a human parsing dataset, which was created specifically for semantic segmentation of multiple parts of human body. We choose this dataset because it provides training data where only some certain parts of human body, such as hand, are available, and these body parts are commonly presented in video actions.

Furthermore, the original LIP dataset contains 19 semantic parts.  Due to the relatively low resolution of the frames, the generated pseudo labels may contain a certain amount of noisy pixel labels for fine-grained human parsing. Thus we merge all 19 human parts into a single human segmentation, which leads to much stronger robustness on segmentation results. Finally, we employ PSPNet \cite{DBLP:conf/cvpr/ZhaoSQWJ17} with DenseNet-121 \cite{DBLP:conf/cvpr/HuangLMW17} as its backbone for human parsing teacher network.

{\bf Scene recognition network}. We use a large-scale scene recognition dataset, Places365 \cite{zhou2017places}, to train the scene recognition teacher network. Places365 contains 365 scene categories. We employ ResNet152 \cite{he2016deep} as the backbone of the teacher network.

\subsection{The Main Networks}
The proposed KINet has three branches - one main branch for action recognition and the other two branches for auxiliary tasks of scene recognition and human parsing guided by the teacher networks. We use Temporal Segment Network (TSN) structure \cite{DBLP:conf/eccv/WangXW0LTG16} as the action recognition branch, where a 2D network is used as backbone. TSN is able to model long-range temporal information by sparsely sampling a number of segments along the whole video, and then average the representation of all segments. In addition, TSN can also be applied in two-stream architecture, with the second stream for modelling motion information by utilizing optical flow. We set the number of segments for training as  $N_{seg}=3$ in our experiments for an efficient training, and also for a fair comparison against previous state-of-the-art methods.

The three branches of KINet share low-level layers in the backbone. There are two reasons for such design. First, low-level features are generalized over three tasks, and sharing features allow the three tasks to be trained more collaboratively with fewer parameters used.
%
%

The higher level layers are three individual branches, not sharing the parameters, but still exchange information through various integration mechanism, which will be introduced in following sections.

\begin{figure}[t]
\centering
\includegraphics[width=0.8\columnwidth]{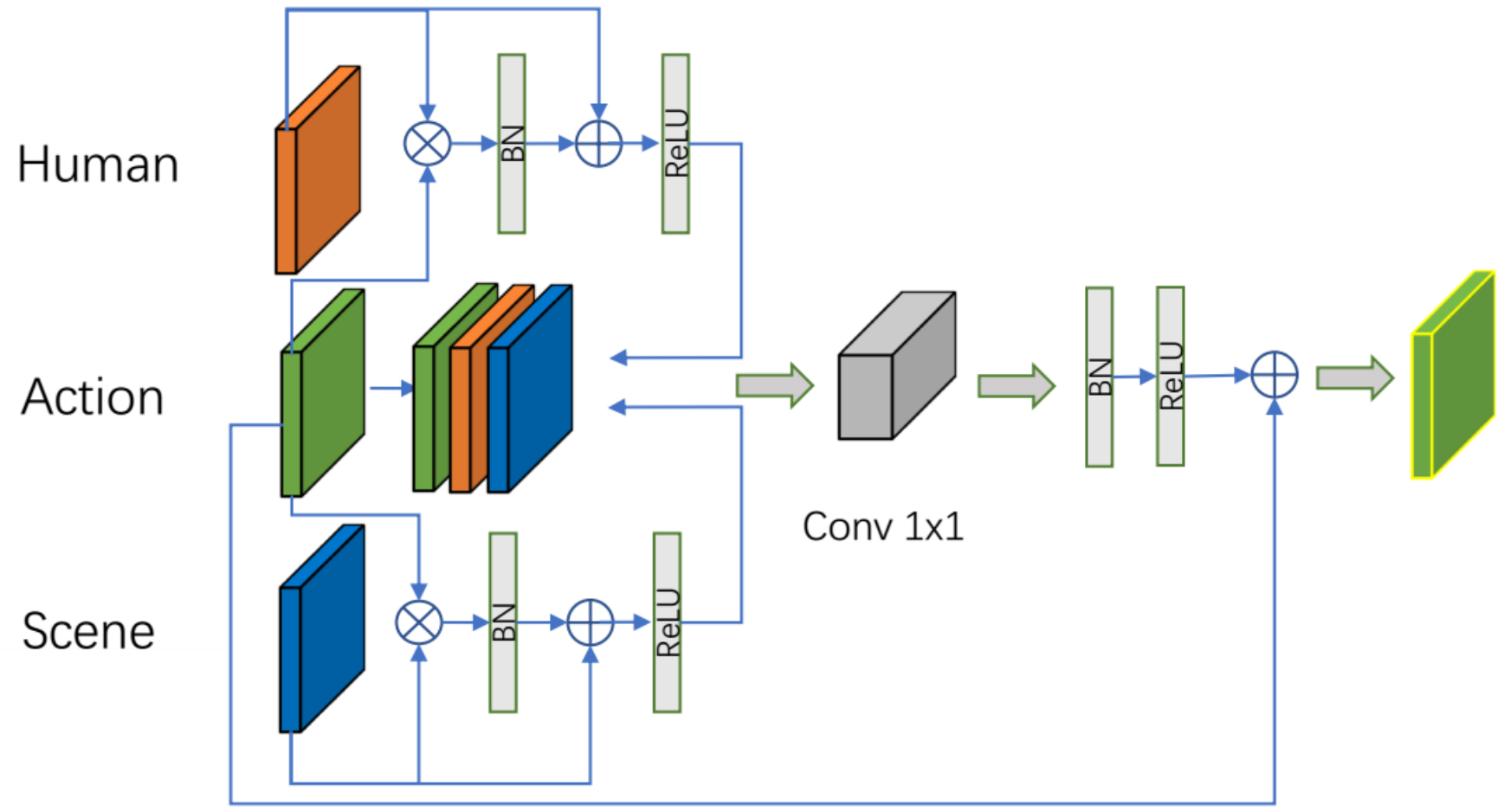} 
\caption{Cross Branch Integration module for aggregating human and scene knowledge into action recognition branch.}
\label{CBI}
\end{figure}

\subsection{Knowledge Integration Mechanism}
Our goal is to develop an efficient feature integration method able to encode context knowledge of different levels.
. 
We propose a two-level Knowledge Integration mechanism including a Cross Branch Integration (CBI) module and an Action Knowledge Graph (AKG) method for encoding the knowledge of human and scene into different levels of the features learned for action recognition.

\subsubsection {Cross Branch Integration (CBI)}\label{cbi section}
The proposed CBI module aims to aggregate the intermediate features learned from two auxiliary branches into action recognition branch, which enables the model to encode the knowledge of human and scene.
Specifically, we use the feature maps of auxiliary branches as gated modulation of the main action features, by implementing element-wise multiplication on them, as shown in Figure \ref{CBI}. We apply a residual-like connection with batch normalization \cite{DBLP:conf/icml/IoffeS15} and ReLU activation \cite{DBLP:conf/icml/NairH10} so that the feature maps of auxiliary branches can directly interact with the action features.

Finally the features maps from three branches are concatenated along the channel dimension, followed by a $ 1 \times 1 $ convolution layer for reducing the number of channels. In this way, the input channel and output channel are guaranteed to be identical, so that the CBI module can be applied at any stage in the network.

\subsubsection {Action Knowledge Graph (AKG)}\label{graph section}


In the final stages,  we apply global average pooling individually on each branch, obtaining three groups of representation vectors with the same size. Each group contains $N_{seg}$ feature vectors, corresponding to the $N_{seg}$ input frames, where $N_{seg}$ means the number of segments in TSN \cite{DBLP:conf/eccv/WangXW0LTG16}. We then construct an Action Knowledge Graph to explicitly model the pair-wise correlation among these representations. To this end, we build an Action Knowledge Graph on the high-level features of three tasks, and apply the recent Graph Convolutional Networks \cite{DBLP:conf/iclr/KipfW17} on the AKG for further integrating high-level semantic knowledge.

\begin{figure}[t]
\centering
\includegraphics[width=1.0\columnwidth]{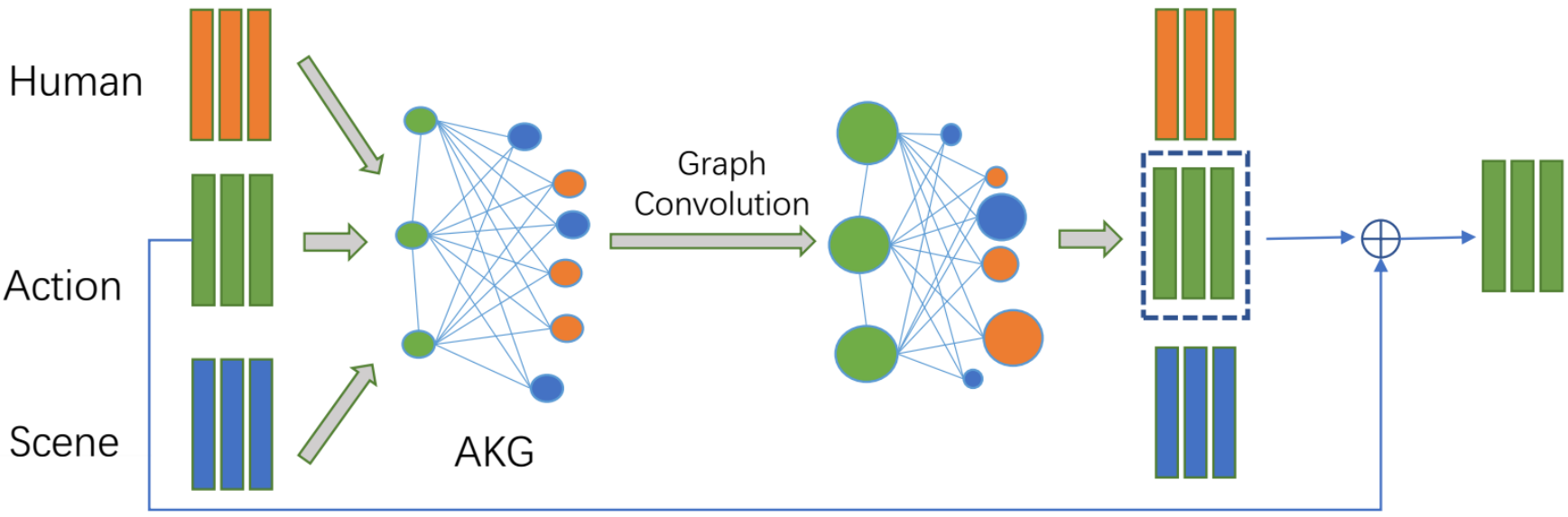} 
\caption{Action Knowledge Graph. We only utilize the outcome of action recognition nodes.}
\label{Graph}
\end{figure}

{\bf Graph Definition.}
 The goal of our knowledge graph is to model the relationship among action, scene and human segments. Specifically, there are $N=3 \times N_{seg}$ total nodes in the graph, denoted as $X=\{x^{action}_i,x^{scene}_i,x^{human}_i|i=1,...,N_{seg}\}$, where the nodes $x^{action}_i,x^{scene}_i,x^{human}_i \in R^{d}$, with $d$ indicating the channel dimension of the last convolutional layer in the backbone. The graph $G \in R^{N \times N}$ represents the pair-wise relationship among the nodes, with edge $G_{ab}$ indicating the relationship between node $x_b$ and node $x_a$.

For the proposed KINet framework, our goal is to build the correlations between the main action recognition and auxiliary scene recognition and human parsing tasks.
Therefore, it is not necessary to construct a fully-connected knowledge graph. We only activate the edges which are directly connected to an action node ${x^{action}_i}$, and set the others to 0. We implement AKG by computing element-wise product of $G$ and an edge mask matrix $I_{mask}$. The mask $I_{mask}$ is a 0 or 1 matrix with the same size as $G$, where the edges between human nodes and scene nodes are set to 0, and otherwise 1.

{\bf Relation function.}
 We describe different forms of the relation function $f$ for computing the relationship between knowledge nodes.

(1) Dot product and embedded dot product. Dot product is a frequently-used function for modelling the similarity between two vectors. It is simple but effective and parameter-free. Another extension of dot product is the embedded dot product, which projects the input vectors onto a subspace, and then applies dot product, by utilizing two learnable weight matrices,

\begin{equation}
f(x_a,x_b)={x_a^Tx_b},
\end{equation}
\begin{equation}
f(x_a,x_b)=\theta(x_a^T)\phi(x_b).
\end{equation}

(2) Concatenation.  We simply adopt the relation module by concatenation which was proposed in  \cite{DBLP:conf/nips/SantoroRBMPBL17}:
\begin{equation}
f(x_a,x_b)=ReLU(W[\theta(x_a),\phi(x_b)]),
\end{equation}
where $[\cdot,\cdot]$ denotes the concatenation operation, and $W$ represents the learnable weight matrix that projects the concatenated vector into a scalar.

{\bf Normalization.} The sum of all edges pointing to the same node must be normalized to 1, and then the graph convolution can be applied to the normalized knowledge graph. In this work, we apply the softmax function for implementing normalization,
\begin{equation}
G_{ab}=\frac {e^{f(x_a,x_b)}}{\sum_{b=1}^Ne^{f(x_a,x_b)}}.
\end{equation}
This normalization function essentially casts dot product into Gaussian function, thus we do not use Gaussian or embedded Gaussian function directly for learning the relations.

\subsection{Relation Reasoning on the Graph}
We apply recent Graph Convolutional Network (GCN) \cite{DBLP:conf/iclr/KipfW17} on the constructed Action Knowledge Graph for aggregating high-level sematic knowledge of human and scene into action recognition branch.
Formally, the behavior of a graph convolution layer can be formulated as,
\begin{equation}
Z=\sigma(I_{mask} \odot GXW),
\end{equation}
\noindent where  $I_{mask}$ is the edge mask matrix mentioned, $G \in R^{N \times N}$ is the matrix of the constructed knowledge graph. $X \in R^{N \times d}$ is the input to the GCN, $W \in R^{d \times d}$ is the learnable weight matrix for GCN, and $\sigma$ is the activation function.
In practice, we found that applying a deeper GCN was not helpful for improving the performance, and thus we only apply one graph convolution layer, which has proved to be enough for modelling rich high-level context information. The output of GCN, $Z \in R^{N \times d}$, has the same size as the input $X$. We only use the $N_{seg}$ vectors from action branch for final classification.

\subsection{Joint Learning}
The proposed three-branch architecture enables an end-to-end joint learning of action recognition, human parsing and scene recognition. The multi-task loss function is computed as,
\begin{equation}
L=\lambda_1L_{action}+\lambda_2L_{human}+\lambda_3L_{scene},
\end{equation}
where $L_{action}$ and $L_{scene}$ are cross-entropy losses for classification, $L_{human}$ is a cross-entropy loss for semantic segmentation. For scene recognition and human parsing, the loss of each segment is calculated individually and then averaged.

The ground truth for action recognition is provided by the training dataset, while the ground truth of scene recognition and human parsing is provided by the two teacher networks as pseudo labels for knowledge distillation. We empirically set $\lambda_1=1$ for main tasks, $\lambda_2=0.01$ and $\lambda_3=0.01$ for the two auxiliary tasks.

\textbf{Learnable parameters for auxiliary tasks}. Notice that previous works, such as \cite{DBLP:conf/cvpr/WuFJS16,DBLP:conf/cvpr/HeilbronBEG17,DBLP:conf/bmvc/WangSWGH16}, encode the extra knowledge by directly using the teacher networks whose parameters are fixed, while our proposed three-branch framework with knowledge distillation enables a joint learning of three tasks. This allows for training three tasks more collaboratively, providing a more principled approach for knowledge integration. Our experiments presented in next section verify this claim.

\section{Experiments}
\subsection{Datasets}
To verify the effectiveness of our KINet, we conduct experiments on a large-scale action recognition dataset Kinetics-400 \cite{DBLP:conf/cvpr/CarreiraZ17}, which contains 400 action categories, with about 240k videos for training and 20k videos for validation. We then examine the generalization ability of our KINet by transferring the learned representation to a small dataset UCF-101 \cite{DBLP:journals/corr/abs-1212-0402}, containing 101 action categories with 13,320 videos in total. Following previous standard criterion, we divide the total videos into 3 training/testing splits and the results of the three splits are averaged as the final result.

\subsection{Implementation Details}

{\bf Training.}
We use ImageNet pretrained weights to initialize the framework. Following the sampling strategy in  \cite{DBLP:conf/eccv/WangXW0LTG16} we uniformly divide the video into $N_{seg}=3$ segments, and randomly select a frame out of each segment. We first resize every frame to size $256 \times 340$ and then we apply multiscale cropping for data augmentation. For Kinetics,
 we utilize SGD optimizer with initial learning rate set to 0.01, which drops by 10 at epoch 20, 40 and 60. The model is totally trained for 70 epochs. We set the weight decay to be $10^{-5}$ and the momentum to be $0.9$. For UCF-101, we follow  \cite{DBLP:conf/eccv/WangXW0LTG16} to fine tune the pretrained weights on Kinetics, where we have all but the first batch normalization layer frozen and the model is trained for 80 epochs.

\noindent{\bf Inference.}
For fair comparison, we also follow  \cite{DBLP:conf/eccv/WangXW0LTG16} by uniformly sampling 25 segments from each video and select one frame out of each segment. We crop the 4 corners and the center of each frame and then flip them so that 10 images are obtained. Totally, there are $25 \times 10=250$ images for each video. We use a sliding window of $N_{seg}=3$ on the 25 test segments. The results are averaged finally to produce the video-level prediction. Note that during inference, the decoder of the human parsing branch and the classifier( fully connected layer) of scene recognition branch can be removed, since our main task is action recognition. This makes it extremely efficient to transfer the learned representation to other datasets.

\begin{table}[tb]
\small
\smallskip
\centering
{
\begin{tabular}{l|l|l|c}
{Method}&{Settings}&  top-1 & gain\\
\hline
\multirow{1}{*}{Baseline}
& TSN-ResNet50 & 69.5 & -\\
\hline
\multirow{3}{*}{KD+Multitask}
&{Baseline+human} & 70.3 & +0.8\\
&{Baseline+scene} & 70.0 & +0.5\\
&{Baseline+human+scene} & 70.6 & +1.1\\

\hline
\multirow{4}{*}{CBI+Multitask}
&{1 CBI$@$res4} & 71.1 & +1.6\\
&{2 CBI$@$res4} & 71.2 & +1.7\\
&{1 CBI$@$res4+1 CBI$@$res5} & 71.8 & +2.3\\
&{2 CBI$@$res4+1 CBI$@$res5} & 71.5 & +2.0\\

\hline
\multirow{3}{*}{AKG+Multitask}
&{AKG + dot product} & 71.7 & +2.2\\
&{AKG + E-dot product} & 71.6 & +2.1\\
&{AKG + concatenation} & 71.2 & +1.7\\
\hline
\multirow{1}{*}{KINet}

&{KINet-ResNet50} & \textbf{72.4} & +2.9

\end{tabular}}
\caption{Ablation study for each components of KINet on Kinetics-400. }
\label{ablation study}
\end{table}

\subsection{Ablation Study on Kinetics}
In this subsection, we conduct extensive experiments on large scale dataset Kinetics to study our framework.
In this study, we use {\bf TSN-ResNet50} \cite{DBLP:conf/eccv/WangXW0LTG16} as baseline.

\begin{table}[tb]
\smallskip
\centering
{
\begin{tabular}{l|l|l}
methods & top-1 & Parameters \\
\hline
TSN-ResNet50 & 69.5 & 24.4M\\

{TSN-ResNet200} & 70.7 & 64.5M\\
\hline
{KINet-ResNet50} & 72.4& 56.9M\\

\end{tabular}
}
\caption{The performance of the proposed KINet, with comparison on parameters.}
\label{combine all}
\end{table}

{\bf Multitask Learning with Knowledge Distillation.} First, in order to show that distilling external knowledge does help with action recognition, we incorporate human parsing and scene recognition into action recognition network, by jointly learning these three tasks via knowledge distillation, yet without applying CBI module or Action Knowledge Graph here. As shown in Table \ref{ablation study}, the multitask learning with knowledge distillation outperforms the baseline. When action recognition and human parsing are jointly trained, the top-1 accuracy increases 0.8\%. When action recognition and scene recognition are jointly trained, the top-1 accuracy increases 0.5\%. When three tasks are jointly trained, the top-1 accuracy increases 1.1\%.

{\bf Cross Branch Integration Module.} Instead of simple multitask learning, we apply CBI Module to enable intermediate feature exchange. As shown in Table \ref{ablation study}, aggregating human and scene knowledge into action branch strengthens the learning ability of action branch. We further employ multiple CBI Modules at different stages, showing that higher accuracy can be obtained. According to experiment results, we finally apply 1 CBI at res4 and 1 CBI at res5 for a balance between accuracy and efficiency.

{\bf Action Knowledge Graph.} The AKG is applied at the late stage of the framework, with 3 possible relation function as mentioned in section \ref{graph section}. We compare their performance in Table\ref{ablation study}. The AKG boosts performance by aggregating multiple branches and models the relation among action, human and scene representations. We find that dot product and embedded dot product are comparable, which are slightly better than ReLU concatenation. We simply choose to use dot product as the relation function in the remaining experiments.

\begin{table}[tb]
\smallskip
\centering
{
\begin{tabular}{l|l|l|l}
Backbones & {TSN top-1} & {KINet top-1} & {Gain}\\
\hline
ResNet50 & 69.5 & 72.4 & {+\textbf{2.9}}\\

BN-Inception & 69.1 & 71.8 & {+\textbf{2.7}}\\
{Inception V3} & 72.5 & 74.1 &{+\textbf{1.6}}\\

\end{tabular}
}
\caption{KINet consistently improve the performance with different backbones.}
\label{backbone table}
\end{table}

\begin{table}[tb]
\smallskip
\centering
{
\begin{tabular}{l|l}
methods  & top-1  \\
\hline
Baseline TSN-ResNet50 & 69.5 \\
\hline

{Fixed auxiliary branches KINet-ResNet50} & 70.5 \\
{Learnable auxiliary branches KInet-ResNet50} & \textbf{72.4}\\

\end{tabular}
}
\caption{Learnable auxiliary branches is better than pre-trained fixed ones for action recognition.}
\label{direct guider table}
\end{table}

 \begin{table*}[!htb]
 \small
\smallskip
\centering

{
\begin{tabular}{l|l|l|l|l}

{}&{Model}& Backbone & top-1 & top-5\\
\hline
\multirow{8}{*}{2D Backbones}
&TSN \cite{DBLP:conf/eccv/WangXW0LTG16} & ResNet50 & 69.5 & 88.9\\
&TSN \cite{DBLP:journals/corr/abs-1710-08011} & ResNet200 & 70.7 & 89.2\\
&TSN \cite{DBLP:conf/eccv/WangXW0LTG16} & BNInception & 69.1 & 88.7\\
&TSN \cite{DBLP:conf/eccv/WangXW0LTG16} & {Inception V3} & 72.5 & 90.2\\
&StNet \cite{DBLP:conf/aaai/HeZGLLLWW19} & ResNet50 & 69.9 & -\\
&StNet \cite{DBLP:conf/aaai/HeZGLLLWW19} & ResNet101 & 71.4 & -\\
&TBN \cite{DBLP:conf/aaai/LiSL019} & ResNet50 & 70.1 & 89.3\\

&Two-stream TSN \cite{DBLP:conf/eccv/WangXW0LTG16} & {Inception V3} & 76.6 &92.4\\
\hline
\multirow{7}{*}{3D Backbones}
&ARTNet  \cite{DBLP:conf/cvpr/WangL0G18} & {ARTNet ResNet18 + TSN} & 70.7 & 89.3\\
&ECO \cite{ECO_eccv18} &{ECO 2D+3D} & 70.0 & 89.4\\
&S3D-G \cite{xie2018rethinking} & {S3D Inception} & 74.7 & 93.4\\
&Nonlocal Network \cite{DBLP:conf/cvpr/0004GGH18} &{I3D ResNet101} & 77.7 & 93.3\\
&SlowFast  \cite{DBLP:journals/corr/abs-1812-03982} &{SlowFast 3D ResNet101+NL} &79.8 & 93.9\\
&I3D \cite{DBLP:conf/cvpr/CarreiraZ17} &{I3D Inception}&71.1 & 89.3\\
&Two-stream I3D \cite{DBLP:conf/cvpr/CarreiraZ17} &{I3D Inception} & 74.2 & 91.3\\

\hline
\multirow{3}{*}{2D Backbones}
&KINet(Ours) & BN-Inception & 71.8 & 89.7\\
&KINet(Ours) & {ResNet50} & 72.4 & 90.3\\
&KINet(Ours) & {Inception V3} & 74.1 & 91.0\\
&{Two-stream KINet(Ours)} & {Inception V3} & 77.8 & 93.1\\

\end{tabular}}
\caption{Comparison with state-of-the-art on Kinetics. }
\label{cmp kinetics stoa}
\end{table*}

\begin{table}[!htb]
\small
\smallskip
\centering

{
\begin{tabular}{l|l|c}

{Model}& Backbone & {top-1}\\
\hline

TS TSN \cite{DBLP:conf/eccv/WangXW0LTG16} & BNInception & 97.0\\
TS TSN \cite{DBLP:conf/eccv/WangXW0LTG16} & {Inception V3} & 97.3\\
TS I3D (Carreira at al. 2017) &{I3D Inception} & 98.0\\
StNet-RGB \cite{DBLP:conf/aaai/HeZGLLLWW19} & ResNet50 & 93.5 \\
StNet-RGB \cite{DBLP:conf/aaai/HeZGLLLWW19} & ResNet101 & 94.3\\

\hline

{TS KINet} & {Inception V3} & 97.8\\

\end{tabular}}
\caption{Comparison with state-of-the-art on UCF-101. ``TS" indicates ``Two-stream". }
\label{cmp ucf-101}
\end{table}

{\bf Entire KINet Framework.} We combine all previous mentioned components into the baseline, $i.e.$ TSN ResNet50, for RGB-based action recognition with entire Knowledge Integration Networks. As shown in Table \ref{ablation study}, we find that the top-1 accuracy has been boosted to 72.4\%, while the baseline is 69.5\%. This significant improvement of 2.9\% on video action recognition benchmark proves the effectiveness of our proposed framework.

{\bf Effective Parameters.} As shown in Table \ref{combine all}, although our method introduces more parameters due to the multi-branch setting, the overall amount of parameters is still less than that of TSN-ResNet200 \cite{DBLP:journals/corr/abs-1710-08011}, yet with higher accuracy. This comparison proves that the functionality of our framework contributes vitally to action recognition, not just because of the extra parameters introduced.

{\bf Different Backbones.} We implement KINet with different backbones, to verify the generalization ability. The results in Table \ref{backbone table} show that our KINet can consistently improve the performance with different backbones.

{\bf Learnable parameters.} To verify the impact of joint learning, we directly use the two teacher networks to provide auxiliary information, with their weights fixed. The results are shown in Table \ref{direct guider table}. The KINet outperforms the fixed way significantly. We explain this phenomenon by stressing the importance of pseudo label guided learning. With KINet, the auxiliary branches are jointly trained with the action recognition branch using the pseudo label, so that the intermediate features of scene and human can be finetuned to suit action recognition better. Yet for the fixed way in previous works, the auxiliary representation cannot be finetuned. Although the fixed auxiliary networks provide more accurate scene recognition and human parsing results compared to the KINet, their improvement on main task, action recognition, is less than that of KINet (1.9\%).

\subsection {Comparison with State-of-the-Art Methods}
Here we compare our 2D KINet with state-of-the-art methods for action recognition, including 2D and 3D methods, on action recognition benchmark Kinetics-400. We also include two-stream CNNs for KINet, where the RGB stream CNN is our KINet and the optical flow stream CNN is normal TSN structure. As shown in Table \ref{cmp kinetics stoa}, our method achieves state-of- the-art results on Kinetics. Although our network is based on 2D backbones, the performance is even on par with state-of- the-art 3D CNN methods.

\subsection {Transfer Learning on UCF-101}
We further transfer the learned representation on Kinetics to smaller dataset UCF-101 to check the generalization ability of our framework.  Following the standard TSN protocol, we report the average of three train/test splits in Table \ref{cmp ucf-101}. The results show that our framework pretrained on Kinetics has strong transfer learning ability. Our model also obtain state-of- the-art result on UCF-101.

\subsection {Visualization}
\begin{figure}[!htb]
\begin{center}
\includegraphics[width=1.0\columnwidth]{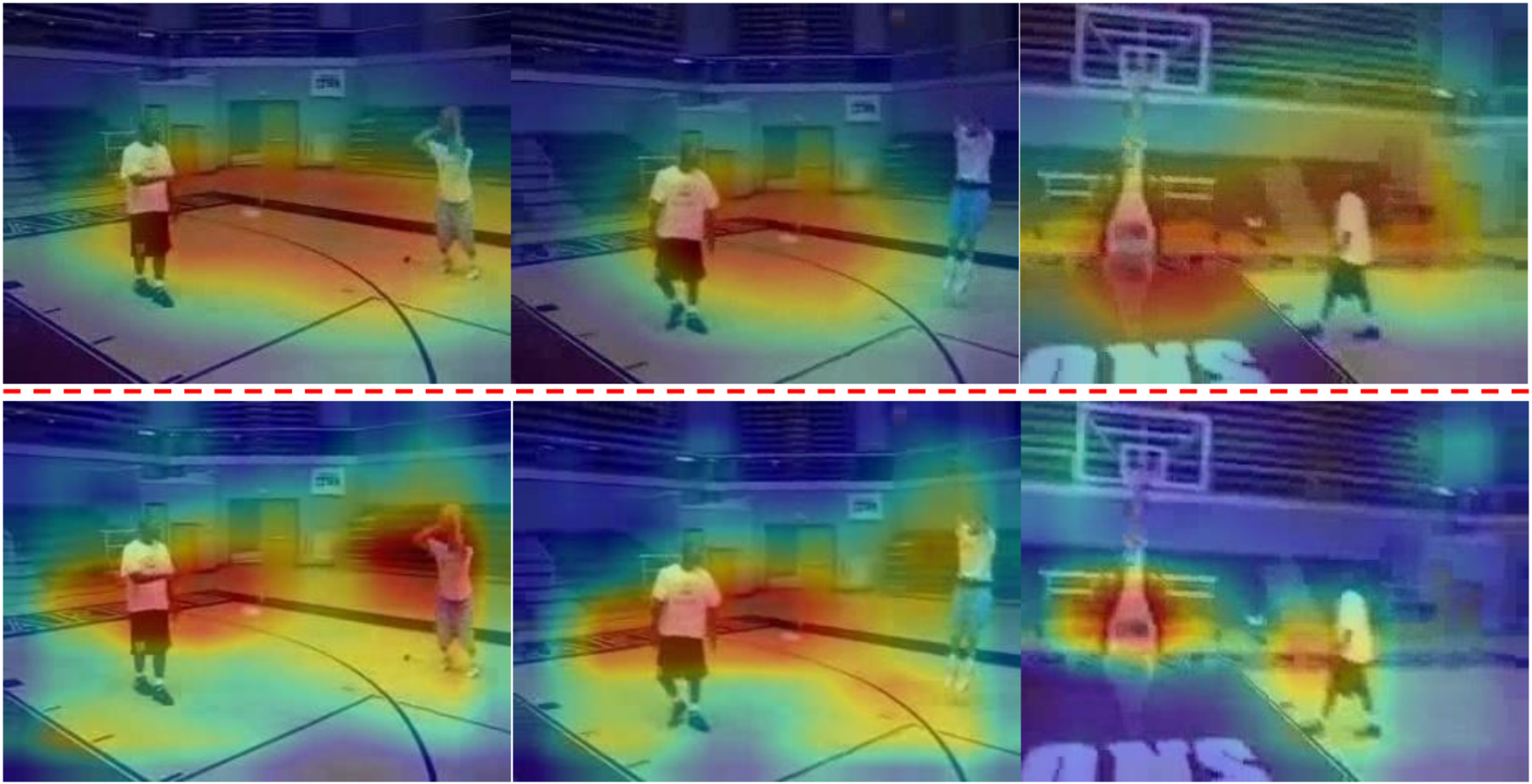}
\end{center}
\caption{Visualization of playing basketball. Top: the activation maps from  TSN-ResNet50, Bottom: the activation maps from KINet-ResNet50.}
\label{visualization}
\end{figure}

As shown in Figure \ref{visualization}, our KINet has more clear understanding of human and scene concepts, compared with baseline TSN. This integration of multiple domain specific knowledge enables our KINet to recognize complex action involving with various high-level semantic cues.

\section{Conclusion}
We have presented new Knowledge Integration Networks (KINet) able to incorporate external semantic cues into action recognition via knowledge distillation. 
Furthermore, a two-level knowledge encoding mechanism is proposed by introducing a Cross Branch Integration (CBI) module for intergrading the extra knowledge into medium-level convolutional features, and an Action Knowledge Graph (AKG) for learning meaningful high-level context information.

\bibliographystyle {aaai}
\bibliography {reference}
\end{document}